\documentclass[letterpaper]{article} 
\usepackage{aaai2026}
\usepackage{times}  
\usepackage{helvet}  
\usepackage{courier}  
\usepackage[hyphens]{url}  
\usepackage{graphicx} 
\usepackage{natbib}  
\usepackage{caption} 
\usepackage{booktabs}%
\usepackage{algorithm}
\usepackage{algorithmic}
\usepackage{multicol}
\usepackage{multirow}
\usepackage{newfloat}
\usepackage{listings}
\DeclareCaptionStyle{ruled}{labelfont=normalfont,labelsep=colon,strut=off} 
\floatstyle{ruled}
\newfloat{listing}{tb}{lst}{}
\floatname{listing}{Listing}
\usepackage{xspace}
\usepackage{xcolor}

\newcommand{\benchmark}{InfoCausalQA\xspace}
\newcommand{\cmark}{\checkmark}       
\newcommand{\xmark}{\ding{55}} 
\usepackage{pifont}
\usepackage{makecell}
\usepackage{amssymb}
\usepackage{amsmath}
\title{\benchmark: Can Models Perform Non-explicit Causal Reasoning Based on Infographic?}
\author{
    Keummin Ka\equalcontrib,
    Junhyeong Park\equalcontrib,
    Jaehyun Jeon,
    Youngjae Yu
}
\affiliations{
    Yonsei University \\
    kummin0429@yonsei.ac.kr, pljh0906@yonsei.ac.kr
}

\usepackage{bibentry}

\begin{document}

\maketitle

\begin{abstract}
Recent advances in Vision-Language Models (VLMs) have demonstrated impressive capabilities in perception and reasoning. However, the ability to perform causal inference—a core aspect of human cognition—remains underexplored, particularly in multimodal settings. In this study, we introduce \benchmark, a novel benchmark designed to evaluate causal reasoning grounded in infographics that combine structured visual data with textual context. The benchmark comprises two tasks: Task 1 focuses on quantitative causal reasoning based on inferred numerical trends, while Task 2 targets semantic causal reasoning involving five types of causal relations—cause, effect, intervention, counterfactual, and temporal. We manually collected 494 infographic–text pairs from four public sources and used GPT-4o to generate 1,482 high-quality multiple-choice QA pairs. These questions were then carefully revised by humans to ensure they cannot be answered based on surface-level cues alone, but instead require genuine visual grounding. Our experimental results reveal that current VLMs exhibit limited capability in computational reasoning and even more pronounced limitations in semantic causal reasoning. Their significantly lower performance compared to humans indicates a substantial gap in leveraging infographic-based information for causal inference. Through \benchmark, we highlight the need for advancing the causal reasoning abilities of multimodal AI systems.
\end{abstract}


\begin{figure}[t]
\centering
\includegraphics[width=0.98\columnwidth]{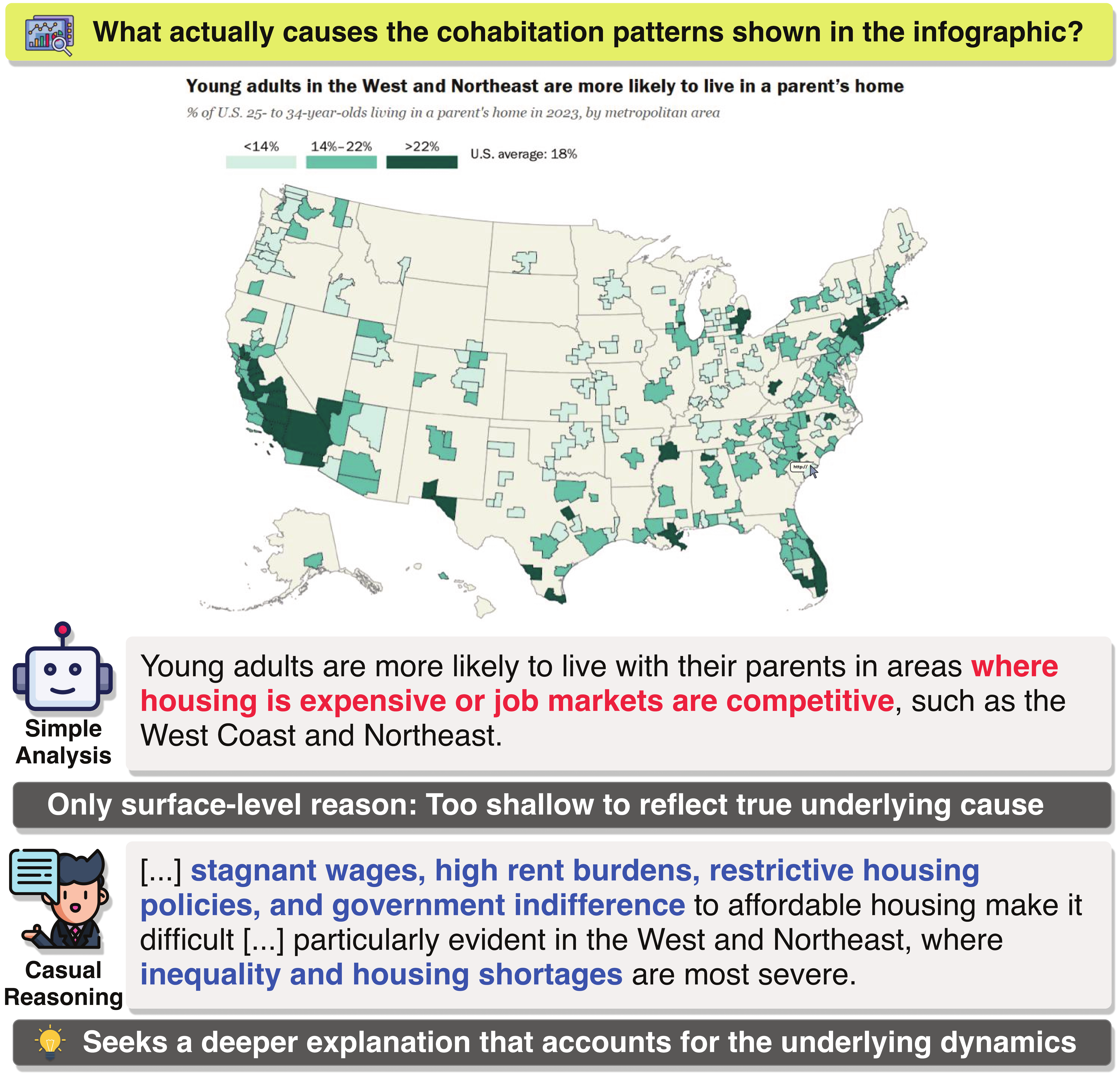}
\caption{A motivating example for causal reasoning in infographics. While traditional infographic analysis often involves simple, superficial inference, causal reasoning requires inferring information that is not explicitly revealed.}
\label{fig:teaser}
\end{figure}

\section{Introduction}
Understanding multiple causal explanations is crucial in real-world scenarios. For example, the sharp decline in stock prices could be attributed to market volatility, but it could also reflect investor panic triggered by policy uncertainty. In policy, science, and everyday decisions, grasping how events influence one another matters more than simply describing them. As a core facet of human cognition, causal reasoning enables interpretation, intention inference, and prediction \cite{pearl2009causality, CLEVRER2020ICLR}. To behave intelligently in complex environments—especially multimodal ones where causal links are often implicit—AI systems must acquire this ability \cite{li2025multimodalcausalreasoningbenchmark}. Systematic evaluation across modalities is essential for robust, explainable decision-making.

\begin{table*}[]
\centering
\small
\begin{tabular}{c|cccc}
\toprule
Benchmark   & \# Samples & Data Types & Causal Reasoning & Main Question Types \\
\midrule
\makecell[c]{ChartQA \\ \cite{chartqa}}  & 32,719     & Bar, Line, Pie Charts & \xmark & \makecell[c]{Compositional question\\ Visual question} \\
\midrule
\makecell[c]{ChartQA-Pro \\ \cite{chartqapro}} & 1,948      & \makecell[c]{Bar, Line, Pie, \\Area, Infographic} & \xmark & \makecell[c]{Mathematical Reasoning \\ Visual Reasoning} \\
\midrule
\makecell[c]{ChartQA-X \\ \cite{chartqax}} & 30,299     & \shortstack{Bar, Line, Pie Charts} & \xmark & \makecell[c]{Descriptive\\ Reasoning} \\
\midrule
\makecell[c]{InfographicVQA \\ \cite{infographicvqa}} & 30,035  & Infographics & \xmark & \makecell[c]{Questions tagged by Evidence \\ Questions tagged by Operation} \\
\midrule
\makecell[c]{InfoChartQA \\ \cite{infochartqa}} & 55,091     & \makecell[c]{Bar, Line, Pie,\\ Infographics} & \xmark & \makecell[c]{Text-based questions\\ Visual-element-based questions} \\
\midrule
\textbf{\makecell[c]{InfoCausalQA \\(Ours)}} & 1,482 & \makecell[c]{Infographics \\(including various charts)} & \cmark & \makecell[c]{Quantitative Causal Reasoning\\ Semantic Causal Reasoning} \\
\bottomrule
\end{tabular}
\caption{Comparison of \benchmark with existing infographic and chart benchmarks}
\label{tab:causal_question_comparison}
\end{table*}

One promising modality for evaluating such reasoning is infographics, which combines visual layouts such as charts with additional textual descriptions. For instance, as illustrated in Figure~\ref{fig:teaser}, the infographic shows regional differences in the rate at which young adults live with their parents. While this may appear as simple geographic data, true causal understanding involves reasoning about latent factors like wage stagnation, housing policy. Infographics thus challenge models to go beyond surface-level interpretation: they must integrate multimodal inputs and infer underlying mechanisms that are not explicitly shown.

Yet, these capabilities remain limited in current multimodal AI systems. There have been numerous cases demonstrating the evolution of visual processing tasks, from early attempts at simple image recognition to more recent efforts to assess the comprehension of infographics.~\cite{9093269, kahou2018, Methani, Kafle_2018_CVPR} While Vision-Language Models(VLMs) are used for analyzing infographics, most benchmarks still neglect causal inference, focusing mainly on surface-level perception and reasoning. While there have been prior attempts to evaluate visual causal reasoning in VLMs, they were limited by a narrow range of task categories~\cite{komanduri2025} or non-reality based generality.~\cite{wang2025} Even most current benchmarks for infographic understanding, as shown in Table~\ref{tab:causal_question_comparison}, focus on simple numerical calculations, direct information retrieval, or basic reasoning over explicitly presented data, without addressing the crucial question~\cite{chartqa, chartqapro, chartqax, infographicvqa, infochartqa}:  \textit{Can models perform non-explicit causal reasoning based on infographic?}

\begin{figure*}[t]
\centering
\includegraphics[width=0.99\textwidth]{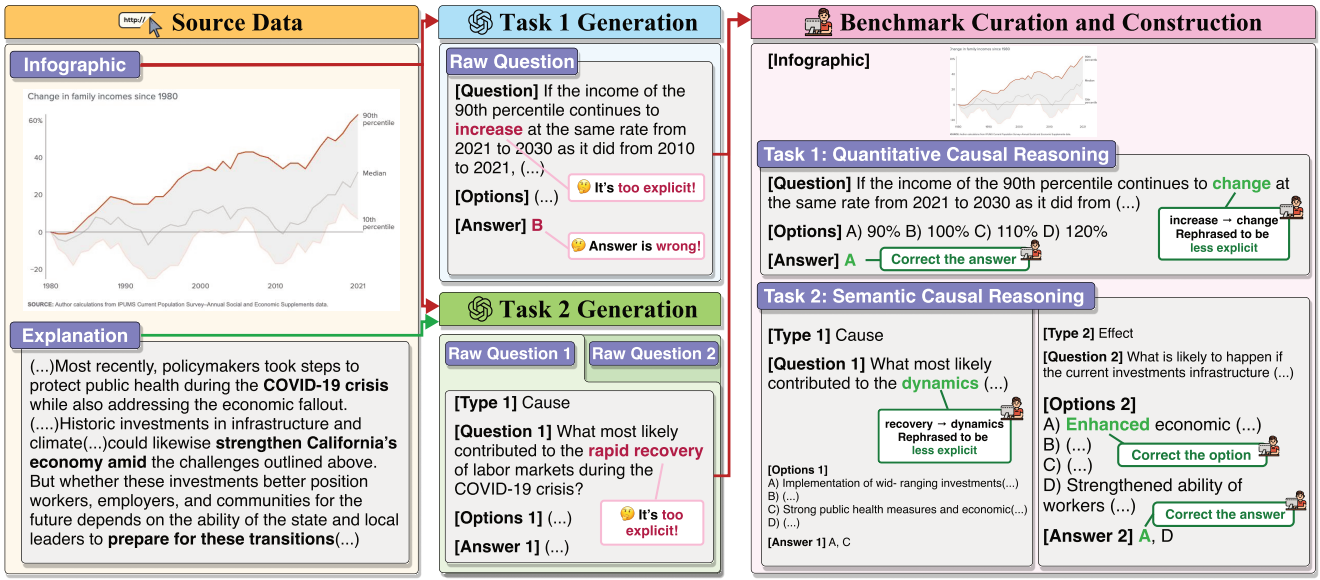}
\caption{An overview of the benchmark generation and construction process. Human annotators manually revised the questions and answer choices generated by GPT-4o to construct the final benchmark.}
\label{fig:pipeline}
\end{figure*}

To address this gap, we introduce \benchmark, a dataset of 1,482 multiple-choice questions curated from 494 infographics. As shown in Figure~\ref{fig:pipeline}, \benchmark comprises two tasks: Task 1, \textit{Quantitative Causal Reasoning}, requires models to infer causal relationships from visual trends, going beyond simple arithmetic or explicit comparisons. For example, in the rightmost Task 1 example of Figure~\ref{fig:pipeline}, the model must predict what happens next if an income trend continues, which involves reasoning about implicit causal dynamics. Task 2, \textit{Semantic Causal Reasoning}, evaluates the model's ability to reason about five core causality types—Cause, Effect, Intervention, Counterfactual, and Temporal—through both visual and textual interpretation. As shown in the rightmost Task 2 example of Figure~\ref{fig:pipeline}, the model may be asked to identify the most likely cause of an income trend or infer the expected effect of a policy change. Such questions require contextual reasoning beyond surface-level visual understanding, relying on inference over implicit causal structures.

We evaluate a range of VLMs on this benchmark. Our experimental results reveal that, despite acceptable perceptual and linguistic capabilities, VLMs struggle with both causal reasoning tasks. These results suggest a fundamental gap between perception and causal inference, underscoring the need for better causal reasoning in models. We hope \benchmark fosters research bridging perception and causal reasoning, toward more explainable multimodal AI.

Our main contributions are as follows:
\begin{itemize}
    \item We propose \benchmark, the first benchmark specifically designed to evaluate causal reasoning over infographics by leveraging rich visual-linguistic information, including relationships not explicitly observable.
    \item We conduct a systematic analysis of current VLMs’ performance on infographic-based causal reasoning tasks, revealing critical limitations in their inference capabilities.
    \item Our study offers insights into future directions for developing inference-driven and explainable AI systems, emphasizing the importance of causal understanding in multimodal reasoning.
\end{itemize}

\section{Related Works}
\subsubsection{Infographics Question Answering Benchmark}

Recent efforts have explored building benchmarks for question answering on infographics and charts with VLMs. \citet{chartqa} introduced ChartQA, a large-scale chart QA benchmark requiring compositional and visual reasoning over bar, line, and pie charts. Its follow-ups, ChartQA-Pro \cite{chartqapro} and ChartQA-X \cite{chartqax}, broadened the scope by including additional chart types and question formats, yet they remained oriented towards mathematical or descriptive reasoning tasks, without evaluating causality. \citet{infographicvqa} targets rich infographic images, requiring joint text–visual understanding and categorizing questions by evidence source and operation type; however, its queries remain largely confined to retrieving information or performing basic arithmetic. \citet{infochartqa} further scales up on paired plain and infographic charts, introducing text-based and visual-element-specific queries to test design-driven comprehension, yet it similarly omits causal questions. As a result, none of these benchmarks probe a model’s ability to perform causal inference over multimodal (chart+text) information. \benchmark fills this gap by introducing questions explicitly focused on causal reasoning: both quantitative causal analysis of chart trends and semantic causal reasoning grounded in infographic-style data.

\subsubsection{Casual Inference}

Interest in how AI systems understand and reason about causality has surged in recent years, giving rise to a series of focused benchmarks. Early efforts such as the structured‐data suite of \citet{cai2023} and the text‐centric CLadder~\cite{jin2023}, CausalBench~\cite{wang2024}, and CausalNet \citep{ashwani2024} probe large language models with hypothetical or narrative questions that hinge on implicit causal links. Building on this line of work, \citet{imam2025} introduce TemporalVQA, which requires models to infer temporal–causal relations across multiple images, illustrating the need for multimodal reasoning beyond text alone. \benchmark advances this agenda by asking: Can models derive causality from the richly structured, visually dense world of infographics? By combining textual, graphical, and spatial signals in a single task, InfoCausalQA challenges models to integrate heterogeneous cues when tracing cause–effect chains.

\section{Design and Construction of \benchmark}
\benchmark is designed to systematically evaluate causal reasoning capabilities grounded in infographics. It aims to assess whether models can move beyond perceptual recognition to infer causal relationships from both visual and textual elements.

\subsection{Task Definition}
We define the two core tasks to systematically evaluate the ability of Vision-Language Models (VLMs) to perform complex causal reasoning in the context of infographics.

\subsubsection{Task 1: Quantitative Causal Reasoning}
This task evaluates a model’s ability to perform causal inference grounded in numerical data presented within infographics. Unlike traditional chart-based QA tasks that focus on directly extracting, comparing, or computing visible values, it requires models to answer causal questions based on hypothetical changes or interventions. Solving these problems involves recognizing key visual elements, understanding the question's context, and performing relevant numerical reasoning. This task assesses whether VLMs possess a foundational understanding of infographics, including both basic quantitative skills and causal inference over visualized data.

\subsubsection{Task 2: Semantic Causal Reasoning}
\label{sec:task1_definition}
This task is designed to evaluate higher-level causal reasoning that extends beyond numerical changes. It involves inference over semantic relationships, temporal dynamics, and hypothetical scenarios that are often implied but not explicitly stated in the infographic. Here, we define five types of causality: \textit{Effect, Cause, Intervention, Counterfactual, and Intervention.} By defining these types, we enable multifaceted causal reasoning beyond simply predicting trends after infographics. Detailed definitions of each type are provided in Table~\ref{tab:causality_type_definition}.

\begin{table*}[t]
\centering  
\small
\begin{tabular}{l|l}
\toprule
Causality Type & Definition \\
\midrule
Effect & Inferring about possible outcomes or situations following the situation described\\
& or depicted in the infographic.\\
\midrule
Cause & Inferring the reasons or driving forces behind the observed trends or patterns in the infographic.\\
\midrule
Intervention & Infer how the infographic trend or pattern would change if a situation (action or policy) \\
& that did not occur occurred.\\
\midrule
Counterfactual & Infer how the infographic trend or pattern would have been different in a hypothetical scenario  \\
& where a specific situation (such as a number or action) was different. \\
\midrule
Temporal & Infer the reason why the infographic trend or pattern changes over time.\\
\bottomrule
\end{tabular}
\caption{Causality type definition in \benchmark's Task 2}
\label{tab:causality_type_definition}
\end{table*}

\subsection{Benchmark Construction}
\subsubsection{Infographic Source Collection}
We  collect infographic data from the following four reliable sources: Gallup\footnote{https://www.gallup.com}, Our World in Data (Owid)\footnote{https://ourworldindata.org/}, Pew Research Center (Pew)\footnote{https://www.pewresearch.org/}, Public Policy Institute of California (PPIC)\footnote{https://www.ppic.org/}. In addition to the infographic images, we manually collect accompanying explanatory text or related articles for each infographic. These contextual materials are used only in Task 2, where they provide richer background information to support the generation of natural and in-depth causal inference questions. In contrast, Task 1 relies solely on the visual content of the infographic to formulate questions based on numerical trends. All infographic–text pairs are manually curated to ensure high informational quality and relevance, resulting in a final set of 494 carefully selected infographics.

\subsubsection{Question and Answer Set Generation}
Using the collected infographics and associated explanatory texts, we construct QA sets for both tasks with GPT-4o, followed by thorough human refinement. For Task 1, the model receives only the infographic image and generates a numerical causal inference question with four answer choices based on visual trends; one option is correct. Questions are manually revised to ensure clarity, numerical accuracy, and alignment with non-explicit causal reasoning, while avoiding surface-level cues such as directly observable values. For Task 2, GPT-4o takes both the infographic and its accompanying textual description as input to generate questions targeting two of five causality types. Multiple options may be correct. Human annotators refine all outputs, validating question type, answer logic, and distractor plausibility, while ensuring that solving the question requires interpreting the visual content—not just the text. The full prompts are provided in Appendix for transparency and reproducibility.

\subsection{Benchmark Statistics}
\subsubsection{Diversity of Infographic Types} 
\begin{figure}[t]
\centering
\includegraphics[width=0.99\columnwidth]{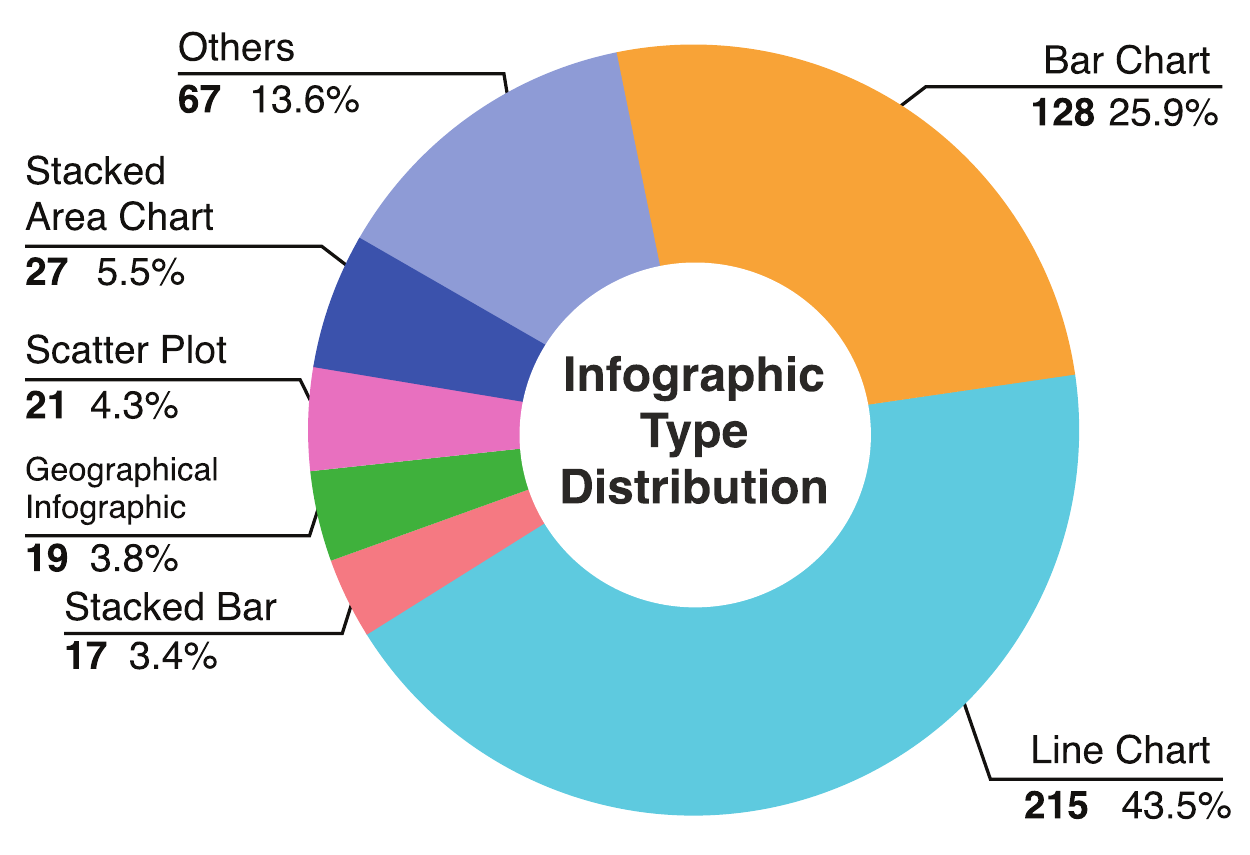}
\caption{Infographics types distribution of \benchmark.}
\label{fig:info_dist}
\end{figure}

To ensure the comprehensiveness and robustness of infographic categories, we analyze the statistical properties and diversity of our dataset. The collection of 494 infographics spans 32 distinct chart types, reflecting a wide range of visualization styles. The majority consists of line charts (43.5\%) and bar charts (25.9\%), which are commonly used for presenting trends and comparisons. Additional, less frequent chart types—used to assess model adaptability to more complex or uncommon formats—are detailed in Figure~\ref{fig:info_dist}.

\begin{figure}[t]
\centering
\includegraphics[width=0.99\columnwidth]{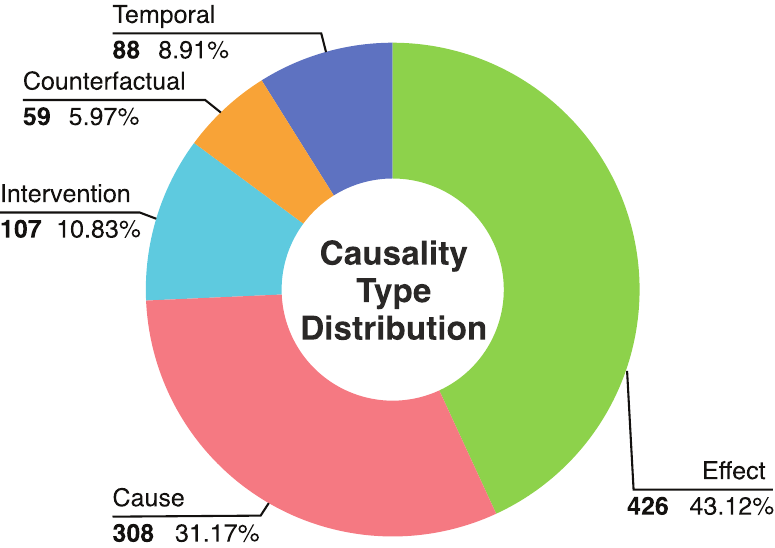}
\caption{Causality type distribution in \benchmark's Task 2.}
\label{fig:type}
\end{figure}

\subsubsection{QA and Causality Type Distribution}
Based on the 494 collected infographics, we construct a total of 1,482 QA instances in \benchmark. Task 1 (Quantitative Causal Reasoning) includes one QA set per single infographic, yielding 494 QA sets. Task 2 (Semantic Causal Reasoning) includes two QA sets per infographic, each targeting a distinct causal reasoning type, resulting in 988 QA sets. Each Task 2 instance is labeled with one of five predefined causality types: Cause, Effect, Intervention, Counterfactual, or Temporal. Among these, \textit{Effect} accounts for the largest proportion (43\%), followed by \textit{Cause} (31\%). This distribution reflects the relative ease with which these causality types can be inferred from infographics and their accompanying textual explanations. The full breakdown is shown in Figure~\ref{fig:type}.

\section{Experiments}
\subsubsection{Models}
We evaluate a diverse set of widely used Vision-Language Models (VLMs), spanning both closed- and open-source models. The closed-source models include o1 \cite{o1}, GPT-4o \cite{gpt4o}, and Claude 4 Sonnet \cite{claude4}, while the open-source models include InternVL3-38B-Instruct \cite{internvl3}, InternVL-2.5-MPO (26B, 8B) \cite{internvl25}, Qwen2.5-VL-Instruct (72B, 32B, 7B) \cite{qwen}, LLaVA-OneVision-7B \cite{onevision}, Idefics2-8B \cite{idefics}, and Phi-3.5-Vision-Instruct \cite{phi}. By covering a broad range of model families and sizes, we aim to provide a comprehensive comparison of current VLMs’ causal reasoning capabilities on infographics. All experiments were conducted on eight RTX 3090 GPUs, each with 24GB VRAM.

\subsection{Quantitative Causal Reasoning}
\subsubsection{Metric}
The model selects the most appropriate one among the four given options and measures the accuracy by comparing it with the correct answer. The full prompts are provided in Appendix for transparency and reproducibility.

\subsubsection{Results}
As shown in Table~\ref{tab:task1_result}, even the strongest models exhibited limited performance on Task 1, with Claude Sonnet 4 achieving the highest accuracy at just 57.3\%. This highlights that, despite their strong language and general reasoning capabilities, current VLMs struggle with complex quantitative causal reasoning that goes beyond surface-level numerical interpretation.

Closed-source models such as Claude, GPT-4o, and o1 generally outperformed open-source counterparts like InternVL and Qwen. However, the large open-source model Qwen2.5-VL-72B-Instruct (52.0\%) achieved performance comparable to GPT-4o, suggesting that open-source models are increasingly competitive.

Among open models, Qwen2.5-VL-72B-Instruct led the group, followed by InternVL3-38B-Instruct (45.7\%). In contrast, smaller models such as Idefics2-8B and Phi-3.5-Vision-Instruct performed substantially worse, with Idefics2-8B ranking the lowest overall. These results suggest that while model scale contributes to performance, size alone is not sufficient for mastering quantitative causal reasoning in infographics.

\subsection{Semantic Causal Reasoning}
\subsubsection{Metric}
In Task 2, models judge each of the four options independently as correct ('O') or incorrect ('X'). This design prevents models from relying on relative comparisons between options—common in fixed-number or "select all that apply" settings—and instead promotes evaluation based solely on the relationship between each option, the question, and the infographic. The full prompts are provided in Appendix for transparency and reproducibility.

We evaluate model performance using metrics of Select-All-That-Apply (SATA) benchmark~\cite{sata}, which allow for precise quantification of multi-option reasoning accuracy. A total of two categories of metrics, Performance and Count bias are used in this paper, and we use the metrics belonging to Performance as main metrics. The metrics belonging to Peformance and their meanings are as follows:
\textit{Exact Match (EM)}: The proportion of questions for which the model selected all correct options without any incorrect ones.
\textit{Precision}: The proportion of selected options that are actually correct.
\textit{Recall}: The proportion of all correct options that were successfully selected by the model.
\textit{Jaccard Index (JI)}: The size of the intersection between the predicted and ground-truth answer sets divided by the size of their union.

\begin{table}[]
\centering  
\begin{tabular}{l|c}
\toprule
Model & Correct Rate (\%) \\
\midrule
Claude Sonnet 4     & 57.3 \\
GPT-4o              & 52.0  \\
o1                 & 53.2  \\
\midrule
InternVL3-38B-Instruct      & 45.7  \\
InternVL2.5-26B-MPO    & 44.1 \\
InternVL2.5-8B-MPO     & 40.7 \\
Qwen2.5-VL-72B-Instruct        & 52.0 \\
Qwen2.5-VL-32B-Instruct        & 49.2 \\
Qwen2.5-VL-7B-Instruct         & 41.7 \\
Llava-Onevision (7B)    & 34.6  \\
Idefics2-8B        & 32.2  \\
Phi3.5-Vision-Instruct (4B)      & 33.8  \\       
\bottomrule
\end{tabular}
\caption{Accuracy evaluation for multiple-choice questions in the quantitative causal reasoning task.}
\label{tab:task1_result}
\end{table}

\subsubsection{Results}
Table~\ref{tab:task2_main_result} shows the overall results for Task 2. We focus on metrics in the Performance category here, while Count Bias metrics are analyzed in Section~\ref{sec:task2_detail}.

Among all models, o1 achieves the highest Exact Match (EM) score (65.18\%), as well as the highest Precision (90.58\%) and Jaccard Index (81.90\%). This indicates that even when it fails to select the exact full set of correct options, it tends to include most correct ones while avoiding incorrect answers. GPT-4o follows with an EM of 56.98\%, but shows higher Recall (87.69\%) than Precision (85.74\%), and a lower Jaccard Index than o1. This suggests that GPT-4o tends to include more false positives, resulting in a slightly less precise prediction set.

Among open-source models, Qwen2.5-VL-72B-Instruct is the only one performing on par with GPT-4o. Most others perform considerably worse. Idefics2-8B ranks lowest in both EM and Jaccard Index (41.68\%), indicating frequent failure to identify correct sets, particularly in multi-answer cases. These results highlight the current limitations of open-source models in accurately handling multi-option causal reasoning.

\begin{table*}[]
\centering
\begin{tabular}{l|ccccccc}
\toprule
\multirow{2}{*}{Model} & \multicolumn{4}{c}{Performance (\%)}  & \multicolumn{3}{c}{Count Bias} \\
\cmidrule(lr){2-5} \cmidrule(lr){6-8} 
& EM $\uparrow$    & Precision $\uparrow$   & Recall $\uparrow$   & JI $\uparrow$  & CtDif   & CtDifAbs $\downarrow$   & CtAcc $\uparrow$     \\
\midrule

Claude Sonnet 4     & 42.81     & 80.41     & 81.65     & 71.31           & 0.04      & 0.56      & 0.51    \\

GPT-4o              & 56.98     & 85.74     & 87.69     & 79.02           & 0.09      & 0.41      & 0.64    \\

o1                     & 65.18     & 90.58         & 86.01      & 81.90             & -0.10       & 0.30          & 0.71       \\
\midrule
InternVL3-38B-Instruct       & 30.67     & 70.52     & 76.11     & 62.28           & 0.17      & 0.57      & 0.50    \\

InternVL2.5-26B-MPO     & 8.00      & 57.14     & 72.83     & 50.81           & 0.47      & 0.67      & 0.40    \\

InternVL2.5-8B-MPO      & 7.29      & 53.33     & 62.10     & 44.06           & 0.22      & 0.64      & 0.44    \\

Qwen2.5-VL-72B-Instruct         & 55.77     & 85.19     & 84.73     & 77.16           & 0         & 0.40      & 0.64    \\

Qwen2.5-VL-32B-Instruct         & 15.69     & 65.49     & 76.59     & 57.70           & 0.34      & 0.62      & 0.41    \\

Qwen2.5-VL-7B-Instruct          & 7.39      & 56.59     & 74.73     & 50.03           & 0.66      & 0.81      & 0.36    \\

Llava-Onevision (7B)  & 24.09     & 74.29     & 76.40     & 60.86           & 0.21      & 0.74      & 0.35    \\

Idefics2-8B         & 4.86      & 51.57     & 64.84     & 41.68           & 0.65      & 0.86      & 0.21   \\

Phi3.5-Vision-Instruct (4B)      & 17.61     & 75.08     & 51.21     & 47.84           & -0.76     & 0.84      & 0.32   \\
\bottomrule
\end{tabular}
\caption{Performance and count bias metrics for the Semantic Causal Reasoning(Task 2).}
\label{tab:task2_main_result}
\end{table*}
\section{Analyses}
In this section, we analyze the Count Bias results from Task 2 and the answers of models according to the causality type, and compare the performance between models and humans through human evaluation.

\subsection{Qualitative Analysis}
To examine the models’ qualitative causal reasoning abilities of Task 2, we analyzed their responses and justifications to understand the causal reasoning pathways used. Figure~\ref{fig:quali} presents comparative results from o1, GPT-4o, and Qwen2.5-VL-72B-Instruct.

\subsubsection{Analyses through Qualitative Examples} 
\begin{figure*}[t]
  \centering
  \includegraphics[width=0.98\textwidth]{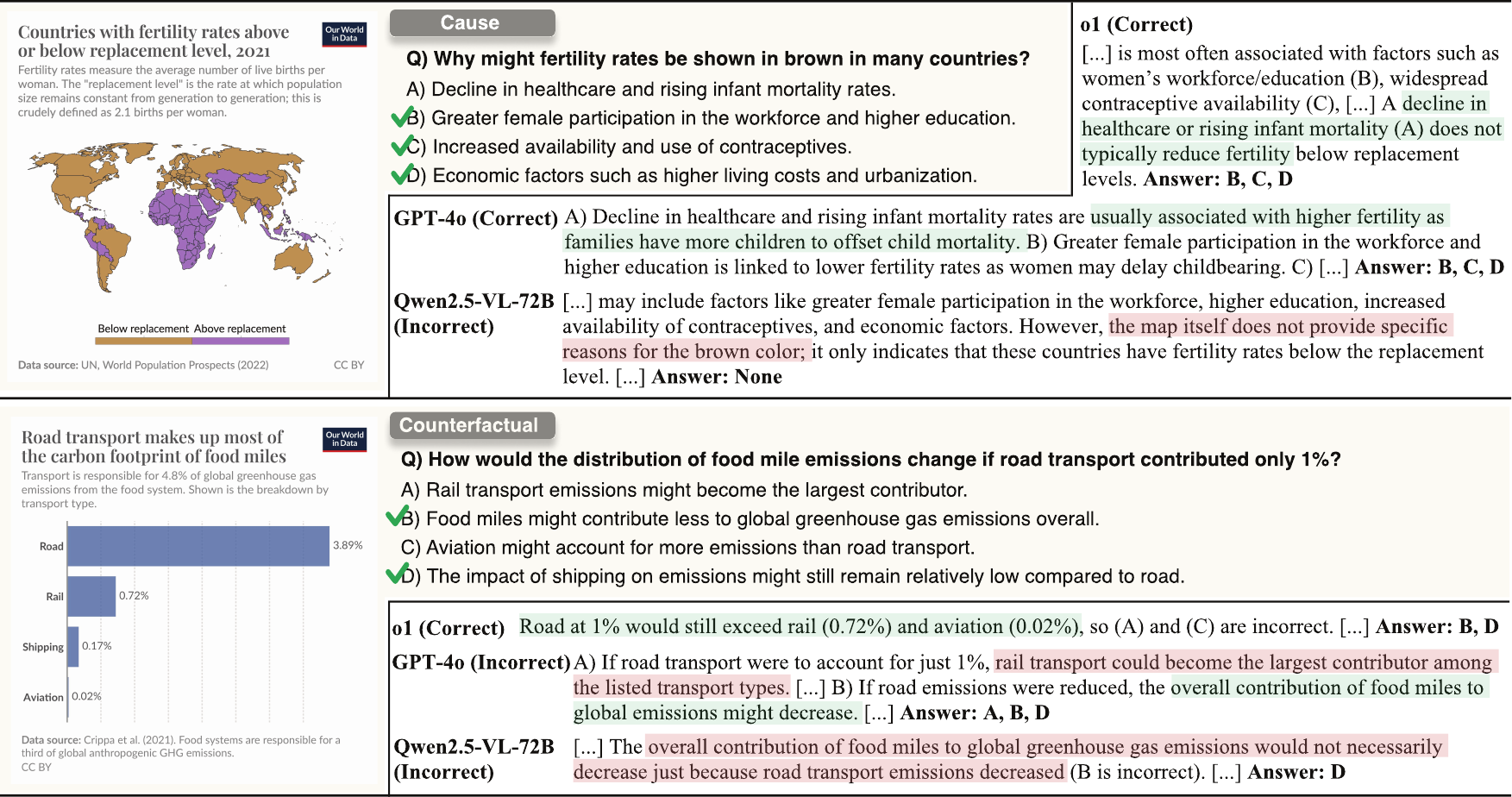}
  \caption{Qualitative results produced by o1, GPT-4o, and Qwen2.5-VL-72B}
  \label{fig:quali}
\end{figure*}

In the top row of Figure~\ref{fig:quali}, the problem is categorized as a \textit{Cause} question. The task is to identify plausible reasons for low fertility rates based on the infographic. Both o1 and GPT-4o correctly selected options B, C, and D, while rejecting A by drawing on background knowledge that poor healthcare and high infant mortality lead to higher fertility rates. Their responses demonstrate an ability to integrate prior knowledge with visual information to perform effective causal reasoning. In contrast, Qwen2.5-VL-72B declined to choose an answer, citing the lack of explicit causal information in the infographic. This overly cautious, data-restricted approach highlights its limited flexibility in causal inference.

In the bottom row of Figure~\ref{fig:quali}, the problem falls under the \textit{Counterfactual} category. The task asks how the distribution of food mile emissions would change if road transport accounted for only 1\%. Only o1 answered correctly, noting that even at 1\%, road emissions would still exceed those of rail and aviation, thus eliminating A and C. GPT-4o partially succeeded: it correctly predicted a decrease in overall emissions (B) but incorrectly assumed that rail would become the highest contributor (A), due to a misjudgment in numerical comparison. Qwen2.5-VL-72B incorrectly rejected B, arguing that reducing road emissions wouldn’t necessarily reduce total emissions—a reasoning error that suggests difficulty with basic counterfactual logic.

\subsection{About Count Bias}
\subsubsection{Metrics for Count Bias Analysis}
We use the metrics belonging to Count Bias and their meanings are as follows:
\textit{Count Difference (CtDif)}: The mean difference between the predicted number of correct options and the actual number (positive for over-selection, negative for under-selection).
\textit{Absolute Count Difference (CtDifAbs)}: The mean absolute error in predicting the number of correct answers, regardless of direction.
\textit{Count Accuracy (CtAcc)}: The proportion of questions in which the model predicted the exact number of correct answers.
\subsubsection{Count Bias Analysis for Task 2}
\label{sec:task2_detail}

Among all models, o1 exhibits the most accurate estimation of the number of correct answers, achieving the lowest CtDifAbs and the highest CtAcc. It also shows a slightly negative CtDif of -0.10, indicating a conservative selection strategy—the model tends to under-select rather than over-select. Combined with its high JI, this suggests that even when o1 fails to identify all correct answers, it avoids choosing incorrect ones and selects options carefully.

By contrast, other models show divergent selection behaviors. For instance, Phi3.5-Vision-Instruct displays a large negative CtDif, reflecting a consistent tendency to under-predict the number of correct answers. On the other hand, Qwen2.5-VL-7B-Instruct shows a positive CtDif, frequently selecting more options than necessary. Phi3.5’s conservative approach is further highlighted by its high Precision (75.08\%) and low JI (47.84\%). This indicates that while the model’s selections are often accurate, they typically cover only a subset of the correct answers—leading to high precision but reduced overlap with the full ground-truth set.

\begin{table}[t]
\centering  
\small
\setlength{\tabcolsep}{1mm}
\begin{tabular}{l|ccccc}
\toprule
Model & E & C & I & CF & T \\
\midrule
Claude Sonnet 4     & 50.70  & 31.17   & 50.47 & 54.24  & 28.41\\

GPT-4o              & 64.55  & 51.95  & 59.81 & 52.54 & 37.50 \\

o1                 & 74.88     & 57.79         & 71.03      & 67.80  & 35.23 \\
\midrule
InternVL3-38B-Instruct     & 35.68 & 20.78  & 42.06 & 33.90  & 25.00  \\

InternVL2.5-26B-MPO    & 9.39   & 6.17   & 14.02   & 8.47   & 0      \\

InternVL2.5-8B-MPO     & 7.98   & 6.17   & 12.15  & 6.78  & 2.27   \\

Qwen2.5-VL-72B-Instruct        & 62.68  & 48.38  & 66.36  & 57.63  & 34.09  \\

Qwen2.5-VL-32B-Instruct        & 17.14  & 12.34  & 14.02  & 18.56  & 20.45 \\

Qwen2.5-VL-7B-Instruct         & 8.22   & 7.79   & 7.48    & 3.39   & 4.55 \\

Llava-Onevision (7B)    & 24.88  & 25.32 & 30.84   & 22.03  & 9.09  \\

Idefics2-8B        & 4.93  & 5.84  & 5.61     & 3.39  & 1.14  \\

Phi3.5-Vision-Instruct (4B)      & 18.31  & 17.21  & 18.69    & 15.25  & 15.91  \\       
\bottomrule
\end{tabular}
\caption{Exact Match (EM) scores by causality type in the semantic causal reasoning task: Effect (E), Cause (C), Intervention (I), Counterfactual (CF), and Temporal (T).}
\label{tab:task1_type_result}
\end{table}

\subsection{Analysis for Causality Type}
We analyze model performance on Semantic Causal Reasoning by causality type, revealing consistent trends across models. The result is shown in Table~\ref{tab:task1_type_result}. Most perform better on \textit{Effect} questions than on \textit{Cause} questions, indicating a relative strength in predicting outcomes rather than identifying underlying reasons. For example, o1 achieves 74.88\% on \textit{Effect} but drops to 57.79\% on \textit{Cause}, with Claude Sonnet 4 showing a similar pattern.

Temporal reasoning is the most difficult category for all models. Even the best-performing models, such as GPT-4o, see a steep decline in accuracy—from 64.55\% on \textit{Effect} to just 37.50\% on \textit{Temporal}. This suggests that understanding time-based causal dynamics, often implicit and abstract in infographics, remains a significant challenge.

While smaller open-source models (e.g., Idefics2-8B, Qwen2.5-VL-7B) perform poorly across all types, models like Llava-Onevision and Phi3.5-Vision-Instruct demonstrate that strong causal reasoning is possible even at smaller scales. These results highlight that model size alone does not guarantee causal inference ability, underscoring the importance of training quality, architecture, and reasoning design.

\section{Conclusion}
In this paper, we introduced \benchmark, a novel and comprehensive benchmark designed to evaluate causal reasoning abilities of Vision-Language Models in the context of real-world infographics. By formulating two distinct but complementary tasks—Quantitative and Semantic Causal Reasoning—we move beyond existing benchmarks that focus on superficial perception or basic computations. Instead, we target the core challenge of inferring underlying causal structures from multimodal inputs, including implicit patterns not explicitly stated.

Through systematic evaluation across a range of state-of-the-art models—spanning both open-source and closed-source VLMs—we reveal pronounced limitations in current systems' ability to handle causal reasoning. While some closed-source models like o1 and GPT-4o show relatively higher performance, even these struggle with tasks requiring deeper integration of visual cues and implicit reasoning. 

By surfacing these challenges, \benchmark contributes a new benchmark paradigm that reflects the complexity of real-world reasoning, where understanding causality is essential for decision-making, explanation, and trust. We hope this benchmark not only drives progress in causally aware model development, but also fosters broader research at the intersection of multimodal understanding, structured reasoning, and explainable AI.

\bibliography{aaai2026}

\begin{thebibliography}{29}
\providecommand{\natexlab}[1]{#1}

\bibitem[{Achiam et~al.(2023)Achiam, Adler, Agarwal, Ahmad, Akkaya, Aleman, Almeida, Altenschmidt, Altman, Anadkat et~al.}]{gpt4o}
Achiam, J.; Adler, S.; Agarwal, S.; Ahmad, L.; Akkaya, I.; Aleman, F.~L.; Almeida, D.; Altenschmidt, J.; Altman, S.; Anadkat, S.; et~al. 2023.
\newblock Gpt-4 technical report.
\newblock arXiv:2303.08774.

\bibitem[{Anthropic(2025)}]{claude4}
Anthropic. 2025.
\newblock Introducing Claude 4.
\newblock \url{https://www.anthropic.com/news/claude-4}.
\newblock Accessed: 2025-05-23.

\bibitem[{Ashwani et~al.(2024)Ashwani, Hegde, Mannuru, Jindal, Sengar, Kathala, Banga, Jain, and Chadha}]{ashwani2024}
Ashwani, S.; Hegde, K.; Mannuru, N.~R.; Jindal, M.; Sengar, D.~S.; Kathala, K. C.~R.; Banga, D.; Jain, V.; and Chadha, A. 2024.
\newblock Cause and Effect: Can Large Language Models Truly Understand Causality?
\newblock arXiv:2402.18139.

\bibitem[{Cai, Liu, and Song(2024)}]{cai2023}
Cai, H.; Liu, S.; and Song, R. 2024.
\newblock Is Knowledge All Large Language Models Needed for Causal Reasoning?
\newblock arXiv:2401.00139.

\bibitem[{Chaudhry et~al.(2020)Chaudhry, Shekhar, Gupta, Maneriker, Bansal, and Joshi}]{9093269}
Chaudhry, R.; Shekhar, S.; Gupta, U.; Maneriker, P.; Bansal, P.; and Joshi, A. 2020.
\newblock LEAF-QA: Locate, Encode \& Attend for Figure Question Answering.
\newblock In \emph{2020 IEEE Winter Conference on Applications of Computer Vision (WACV)}, 3501--3510.

\bibitem[{Chen et~al.(2024)Chen, Wang, Cao, Liu, Gao, Cui, Zhu, Ye, Tian, Liu et~al.}]{internvl25}
Chen, Z.; Wang, W.; Cao, Y.; Liu, Y.; Gao, Z.; Cui, E.; Zhu, J.; Ye, S.; Tian, H.; Liu, Z.; et~al. 2024.
\newblock Expanding performance boundaries of open-source multimodal models with model, data, and test-time scaling.
\newblock arXiv:2412.05271.

\bibitem[{Hegde, Fazli, and Seifi(2025)}]{chartqax}
Hegde, S.; Fazli, P.; and Seifi, H. 2025.
\newblock ChartQA-X: Generating Explanations for Charts.
\newblock arXiv:2504.13275.

\bibitem[{Imam, Lyu, and Aji(2025)}]{imam2025}
Imam, M.~F.; Lyu, C.; and Aji, A.~F. 2025.
\newblock Can Multimodal LLMs do Visual Temporal Understanding and Reasoning? The answer is No!
\newblock arXiv:2501.10674.

\bibitem[{Jaech et~al.(2024)Jaech, Kalai, Lerer, Richardson, El-Kishky, Low, Helyar, Madry, Beutel, Carney et~al.}]{o1}
Jaech, A.; Kalai, A.; Lerer, A.; Richardson, A.; El-Kishky, A.; Low, A.; Helyar, A.; Madry, A.; Beutel, A.; Carney, A.; et~al. 2024.
\newblock Openai o1 system card.
\newblock arXiv:2412.16720.

\bibitem[{Jin et~al.(2023)Jin, Chen, Leeb, Gresele, Kamal, Lyu, Blin, Gonzalez~Adauto, Kleiman-Weiner, Sachan et~al.}]{jin2023}
Jin, Z.; Chen, Y.; Leeb, F.; Gresele, L.; Kamal, O.; Lyu, Z.; Blin, K.; Gonzalez~Adauto, F.; Kleiman-Weiner, M.; Sachan, M.; et~al. 2023.
\newblock Cladder: Assessing causal reasoning in language models.
\newblock \emph{Advances in Neural Information Processing Systems}, 36: 31038--31065.

\bibitem[{Kafle et~al.(2018)Kafle, Price, Cohen, and Kanan}]{Kafle_2018_CVPR}
Kafle, K.; Price, B.; Cohen, S.; and Kanan, C. 2018.
\newblock DVQA: Understanding Data Visualizations via Question Answering.
\newblock In \emph{Proceedings of the IEEE Conference on Computer Vision and Pattern Recognition (CVPR)}.

\bibitem[{Kahou et~al.(2018)Kahou, Michalski, Atkinson, Kadar, Trischler, and Bengio}]{kahou2018}
Kahou, S.~E.; Michalski, V.; Atkinson, A.; Kadar, A.; Trischler, A.; and Bengio, Y. 2018.
\newblock FigureQA: An Annotated Figure Dataset for Visual Reasoning.
\newblock arXiv:1710.07300.

\bibitem[{Komanduri, Bhaila, and Wu(2025)}]{komanduri2025}
Komanduri, A.; Bhaila, K.; and Wu, X. 2025.
\newblock CausalVLBench: Benchmarking Visual Causal Reasoning in Large Vision-Language Models.
\newblock arXiv:2506.11034.

\bibitem[{Lauren{\c{c}}on et~al.(2025)Lauren{\c{c}}on, Tronchon, Cord, and Sanh}]{idefics}
Lauren{\c{c}}on, H.; Tronchon, L.; Cord, M.; and Sanh, V. 2025.
\newblock What matters when building vision-language models?
\newblock \emph{Advances in Neural Information Processing Systems}, 37: 87874--87907.

\bibitem[{Li et~al.(2024)Li, Zhang, Guo, Zhang, Li, Zhang, Zhang, Zhang, Li, Liu et~al.}]{onevision}
Li, B.; Zhang, Y.; Guo, D.; Zhang, R.; Li, F.; Zhang, H.; Zhang, K.; Zhang, P.; Li, Y.; Liu, Z.; et~al. 2024.
\newblock Llava-onevision: Easy visual task transfer.
\newblock arXiv:2408.03326.

\bibitem[{Li et~al.(2025)Li, Wang, Liu, Zhang, Ma, Long, and Cai}]{li2025multimodalcausalreasoningbenchmark}
Li, Z.; Wang, H.; Liu, D.; Zhang, C.; Ma, A.; Long, J.; and Cai, W. 2025.
\newblock Multimodal Causal Reasoning Benchmark: Challenging Vision Large Language Models to Discern Causal Links Across Modalities.
\newblock arXiv:2408.08105.

\bibitem[{Lin et~al.(2025)Lin, Xie, Liu, Ye, Chen, and Liu}]{infochartqa}
Lin, M.; Xie, T.; Liu, M.; Ye, Y.; Chen, C.; and Liu, S. 2025.
\newblock InfoChartQA: A Benchmark for Multimodal Question Answering on Infographic Charts.
\newblock arXiv:2505.19028.

\bibitem[{Masry et~al.(2025)Masry, Islam, Ahmed, Bajaj, Kabir, Kartha, Laskar, Rahman, Rahman, Shahmohammadi et~al.}]{chartqapro}
Masry, A.; Islam, M.~S.; Ahmed, M.; Bajaj, A.; Kabir, F.; Kartha, A.; Laskar, M. T.~R.; Rahman, M.; Rahman, S.; Shahmohammadi, M.; et~al. 2025.
\newblock ChartQAPro: A more diverse and challenging benchmark for chart question answering.
\newblock arXiv:2504.05506.

\bibitem[{Masry et~al.(2022)Masry, Long, Tan, Joty, and Hoque}]{chartqa}
Masry, A.; Long, D.~X.; Tan, J.~Q.; Joty, S.; and Hoque, E. 2022.
\newblock Chartqa: A benchmark for question answering about charts with visual and logical reasoning.
\newblock arXiv:2203.10244.

\bibitem[{Mathew et~al.(2022)Mathew, Bagal, Tito, Karatzas, Valveny, and Jawahar}]{infographicvqa}
Mathew, M.; Bagal, V.; Tito, R.; Karatzas, D.; Valveny, E.; and Jawahar, C. 2022.
\newblock Infographicvqa.
\newblock In \emph{Proceedings of the IEEE/CVF Winter Conference on Applications of Computer Vision}, 1697--1706.

\bibitem[{Methani et~al.(2020)Methani, Ganguly, Khapra, and Kumar}]{Methani}
Methani, N.; Ganguly, P.; Khapra, M.~M.; and Kumar, P. 2020.
\newblock PlotQA: Reasoning over Scientific Plots.
\newblock In \emph{Proceedings of the IEEE/CVF Winter Conference on Applications of Computer Vision (WACV)}.

\bibitem[{Microsoft(2024)}]{phi}
Microsoft. 2024.
\newblock Phi-3 Technical Report: A Highly Capable Language Model Locally on Your Phone.
\newblock arXiv:2404.14219.

\bibitem[{Pearl(2009)}]{pearl2009causality}
Pearl, J. 2009.
\newblock \emph{Causality}.
\newblock Cambridge university press.

\bibitem[{Team(2025)}]{qwen}
Team, Q. 2025.
\newblock Qwen2.5-VL Technical Report.
\newblock arXiv:2502.13923.

\bibitem[{Wang(2024)}]{wang2024}
Wang, Z. 2024.
\newblock Causalbench: A comprehensive benchmark for evaluating causal reasoning capabilities of large language models.
\newblock In \emph{Proceedings of the 10th SIGHAN Workshop on Chinese Language Processing (SIGHAN-10)}, 143--151.

\bibitem[{Wang et~al.(2025)Wang, Zhang, Tang, and Wang}]{wang2025}
Wang, Z.; Zhang, S.; Tang, C.; and Wang, K. 2025.
\newblock TimeCausality: Evaluating the Causal Ability in Time Dimension for Vision Language Models.
\newblock arXiv:2505.15435.

\bibitem[{Xu et~al.(2025)Xu, Cui, Fang, Xue, Eckman, and Reddy}]{sata}
Xu, W.; Cui, S.; Fang, X.; Xue, C.; Eckman, S.; and Reddy, C.~K. 2025.
\newblock Sata-bench: Select all that apply benchmark for multiple choice questions.
\newblock arXiv:2506.00643.

\bibitem[{Yi et~al.(2020)Yi, Gan, Li, Kohli, Wu, Torralba, and Tenenbaum}]{CLEVRER2020ICLR}
Yi, K.; Gan, C.; Li, Y.; Kohli, P.; Wu, J.; Torralba, A.; and Tenenbaum, J.~B. 2020.
\newblock {CLEVRER:} Collision Events for Video Representation and Reasoning.
\newblock In \emph{ICLR}.

\bibitem[{Zhu et~al.(2025)Zhu, Wang, Chen, Liu, Ye, Gu, Tian, Duan, Su, Shao et~al.}]{internvl3}
Zhu, J.; Wang, W.; Chen, Z.; Liu, Z.; Ye, S.; Gu, L.; Tian, H.; Duan, Y.; Su, W.; Shao, J.; et~al. 2025.
\newblock Internvl3: Exploring advanced training and test-time recipes for open-source multimodal models.
\newblock arXiv:2504.10479.

\end{thebibliography}

\clearpage

\appendix

\section{Experimental Setup}
\subsection{List of Vision-Language Models Used in Paper}
All the VLMs we used for this paper are as follows.
\begin{itemize}
    \item o1 \cite{o1}
    \item GPT-4o \cite{gpt4o}
    \item Claude 4 Sonnet \cite{claude4}
    \item InternVL3-38B-Instruct \cite{internvl3}
    \item InternVL-2.5-MPO (26B, 8B) \cite{internvl25}
    \item Qwen2.5-VL-Instruct (72B, 32B, 7B) \cite{qwen}
    \item LLaVA-OneVision-7B \cite{onevision}
    \item Idefics2-8B \cite{idefics}
    \item Phi-3.5-Vision-Instruct \cite{phi}
\end{itemize}
\subsection{Temperature Setting}
For all stages of QA generation and evaluation, we use a sampling temperature of 0.2. This relatively low temperature setting is chosen to minimize randomness in model outputs and promote deterministic reasoning. Since both Task 1 and Task 2 in InfoCausalQA involve multi-step causal inference grounded in visual data, high output consistency is essential to maintain logical coherence and avoid distractive variation in question generation.

Using a lower temperature encourages the model to favor the most likely. Preliminary tests with higher temperatures (e.g., 0.7–1.0) resulted in more diverse but often logically inconsistent questions and answer choices, reinforcing our decision to fix the temperature at 0.2.

\section{Data Sources}
To construct a benchmark grounded in authentic, real-world data, all infographics used in InfoCausalQA were collected from reputable public research organizations. These sources regularly publish high-quality visualizations on social, economic, environmental, and political topics. Below, we briefly introduce each of the four major sources used:
\subsection{Gallup\footnote{https://www.gallup.com}}
A global analytics and advice firm that publishes survey-based infographics on public opinion, workplace engagement, wellbeing, and world affairs. Gallup visualizations are grounded in regular polling across diverse populations and often highlight temporal trends or geographic differences.
\subsection{Our World in Data (Owid)\footnote{https://ourworldindata.org/}}
An open-access platform that combines academic research with interactive visualization. OWID covers global issues including climate change, poverty, health, energy, and food security. The site is widely used in both educational and policy contexts for its clean and information-dense visual representations.
\subsection{Pew Research Center (Pew)\footnote{https://www.pewresearch.org/}}
A nonpartisan fact tank based in the U.S. that conducts public opinion polling, demographic research, content analysis, and other data-driven social science research. Pew regularly releases infographics on topics such as technology use, political attitudes, religion, and generational trends.
\subsection{Public Policy Institute of California (PPIC)\footnote{https://www.ppic.org/}}
A nonprofit, nonpartisan think tank focused on issues affecting California, such as education, healthcare, environment, and housing. Their visual reports often provide insight into regional policy debates using clear and data-rich infographics.

These sources ensure that InfoCausalQA reflects a broad spectrum of data types and topics while maintaining factual reliability and visual authenticity. The use of real-world infographics helps to evaluate models under conditions that resemble practical applications in media analysis, policy review, and public communication.

\section{Full Distribution of Infographic Types}
To analyze the structural diversity of InfoCausalQA, we manually categorized all 494 infographics into 32 distinct chart types. Table~\ref{tab:full_infographic_distribution} reports the full distribution, including both common and rare formats.

While the majority of examples are composed of Line Charts (215, 43.5\%) and Bar Charts (128, 25.9\%), the dataset also includes a variety of less frequent but semantically rich formats such as Choropleth Maps, Sankey Diagrams, Treemaps, and Venn Diagrams. These diverse formats ensure that the benchmark tests model robustness across a wide range of real-world infographic representations.

\section{Prompts for Problem Generation and Answering}
To support the construction and evaluation of the InfoCausalQA benchmark, we designed specialized prompting strategies for both the generation and answering of causal reasoning questions grounded in infographics. The prompts are tailored to each of the two benchmark tasks—Task 1: Quantitative Causal Reasoning and Task 2: Semantic Causal Reasoning—and are further adapted to reflect the unique reasoning formats required in each. This appendix documents the complete set of prompts used during dataset creation and experiments.

\subsection{Problem Generation Prompts}
\subsubsection{For Task 1}
\begin{figure*}[t]
\centering
\includegraphics[width=0.99\textwidth]{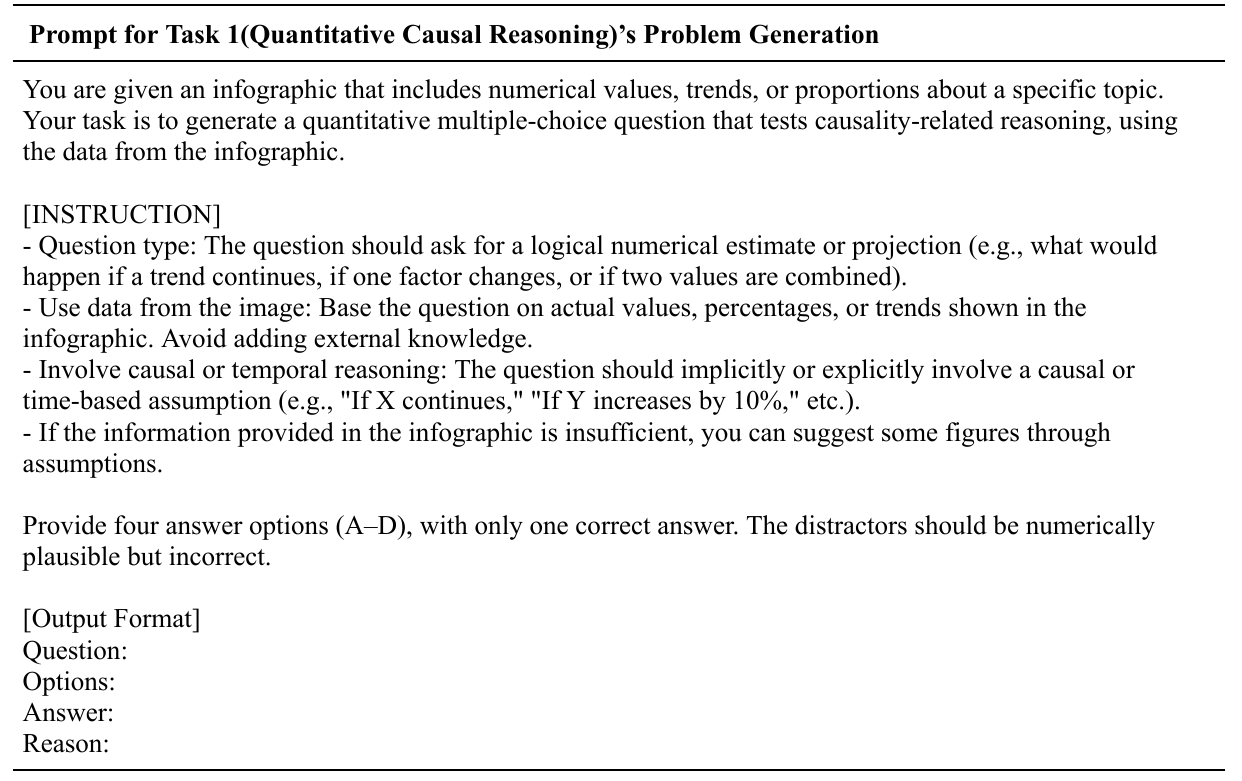}
\caption{Prompt used to generate problems for Task 1(Quantitative Causal Reasoning)}
\label{fig:task1_generation}
\end{figure*}

The goal of Task 1 is to generate questions that test a model's ability to reason causally using numerical information in infographics. Figure~\ref{fig:task1_generation} shows the prompt used to create the initial problem for Task 1. The prompt instructs the model to: 

First, formulate a multiple-choice question (MCQ) based on quantitative data such as trends, percentages, or numerical values present in the infographic. Second, embed causal or temporal inference into the question by including hypothetical scenarios (e.g., “if a trend continues”, “if X increases”). Third, generate four answer options, only one of which is correct, while the others are plausible but incorrect distractors. It also includes the basis for the correct answer so that people can refer to it when making corrections.
This prompt design ensures that questions require causal inference rather than mere data lookup or extraction.
\subsubsection{For Task 2}
\begin{figure*}[t]
\centering
\includegraphics[width=0.99\textwidth]{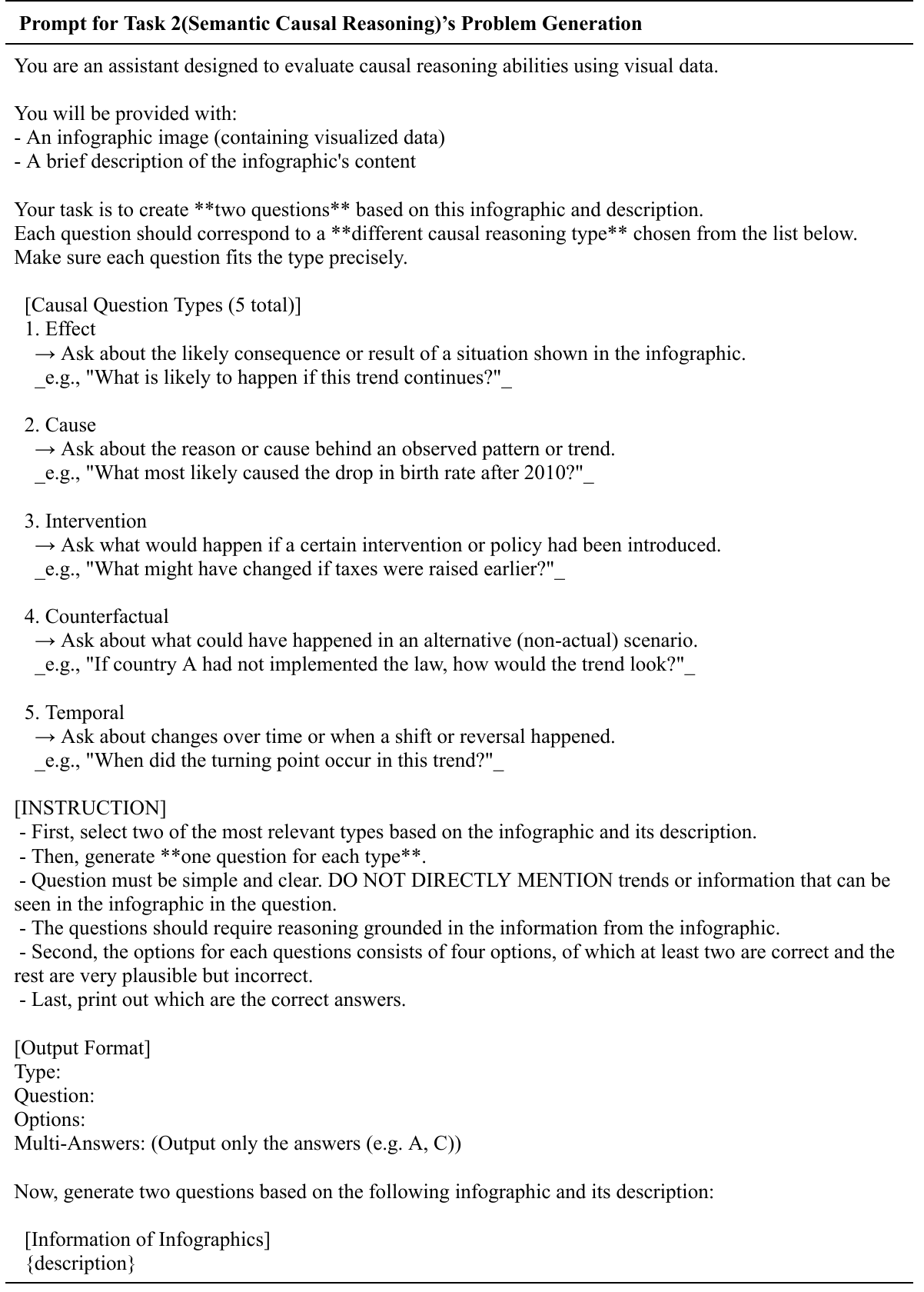}
\caption{Prompt used to generate problems for Task 2(Semantic Causal Reasoning)}
\label{fig:task2_generation}
\end{figure*}

For Task 2, the model is asked to generate questions that evaluate higher-level semantic causal reasoning. Figure~\ref{fig:task2_generation} shows the prompt used to create the initial problem for Task 2.The prompt is provided both an infographic, and an accompanying textual description.

Based on this information, the model must:
First, select two causal reasoning types from a predefined set of five:
Effect, Cause, Intervention, Counterfactual, and Temporal. Second, for each type, generate one question that adheres precisely to its definition. Third, avoid directly quoting or referencing observable trends or specific values from the infographic; instead, questions must rely on reasoning grounded in the visual-textual context. Next, provide four answer choices per question, with at least two correct answers and the rest being plausible but incorrect. And last, clearly indicate which options are correct.
This prompt guides the model to construct complex reasoning questions that simulate analytical interpretation of infographic content.

\subsection{Multiple-Choice Answering Prompts}
To evaluate model performance on the benchmark, we designed two dedicated answering prompts—one for each task—each enforcing strict output formats to facilitate automated scoring.
\subsubsection{For Task 1}
\begin{figure*}[t]
\centering
\includegraphics[width=0.99\textwidth]{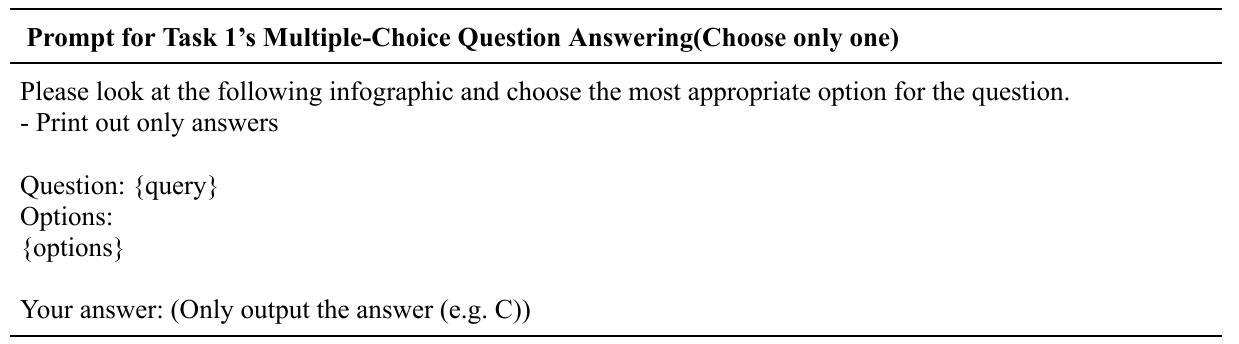}
\caption{Prompt used to solve Task 1(Quantitative Cuasal Reasoning)}
\label{fig:task1_solve}
\end{figure*}

Figure~\ref{fig:task1_solve} shows the prompt used to solve the problem for Task 1. The Task 1 answering prompt presents a single multiple-choice question and instructs the model to:
First, examine the infographic and select the single most appropriate answer (A–D). Then, output only the chosen option letter, without any additional text or explanation. This format is compatible with exact-match accuracy metrics.
\subsubsection{For Task 2}
\begin{figure*}[t]
\centering
\includegraphics[width=0.99\textwidth]{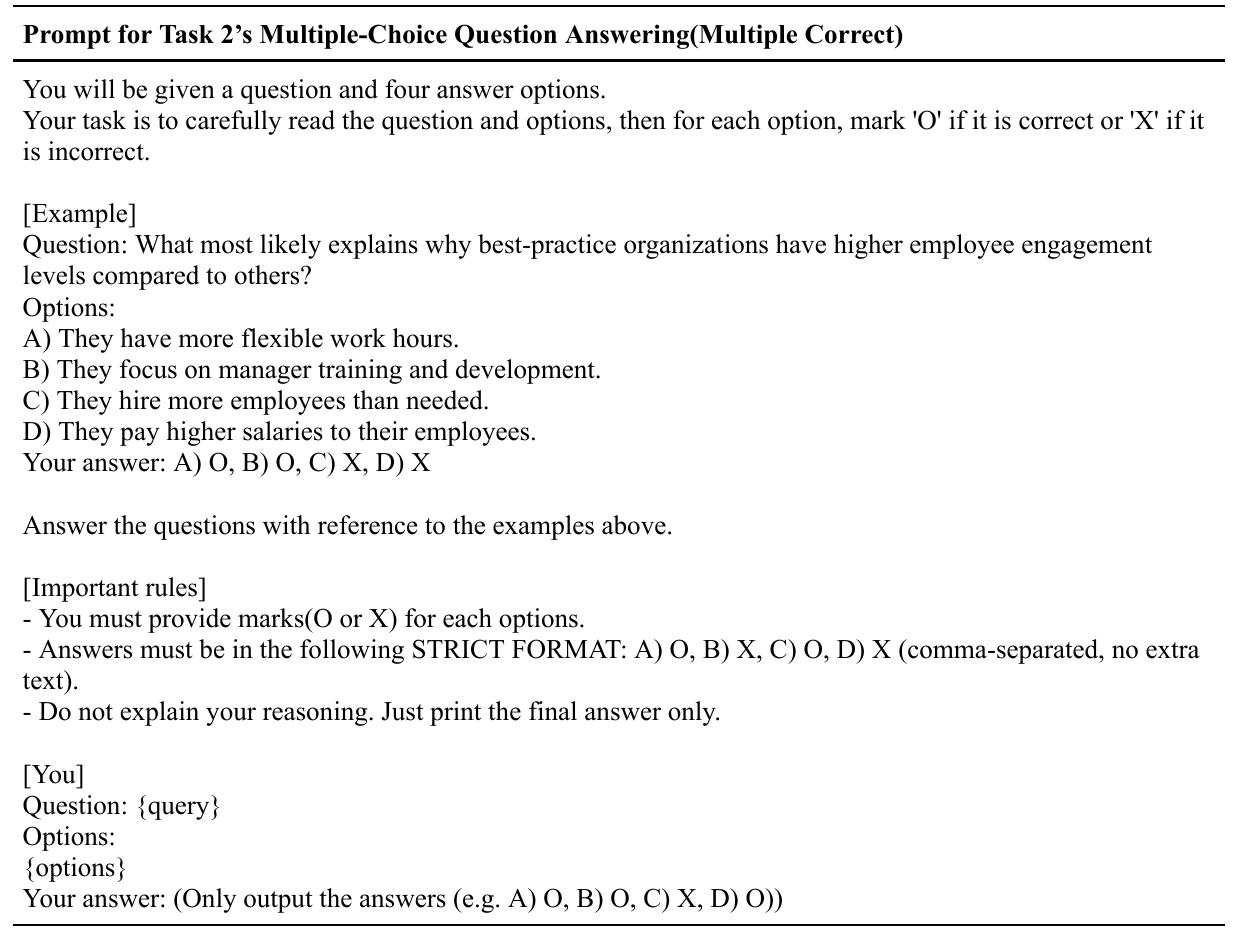}
\caption{Prompt used to solve Task 2(Semantic Causal Reasoning)}
\label{fig:task2_solve}
\end{figure*}

Figure~\ref{fig:task2_solve} shows the prompt used to solve the problem for Task 2. Task 2 requires a multi-answer format. This design prevents models from relying on relative comparisons between options—common in fixed-number or ”select all that apply” settings—and instead promotes evaluation based solely on the relationship between each option, the question, and the infographics. 

The answering prompt provides: First, a semantic causal reasoning question with four answer options. Second, instructions to evaluate each option independently and mark it as either: O (correct), or X (incorrect). The model must output the results in a strict comma-separated format (e.g., A) O, B) X, C) O, D) X). No additional text or justification is allowed. This format allows for partial-credit scoring using multi-label evaluation metrics such as Precision, Recall, and Jaccard Index.

\section{Metrics Definition for Task 2}
We use metrics from SATA-Bench~\cite{sata} to evaluate the correct answer to Task 2. This paper uses a total of seven metrics belonging to the Performance and Count Bias categories. The definitions of each are detailed below.
\subsection{Performance Metrics}
\begin{itemize}
    \item EM(Exact Match): EM measures the proportion of questions for which the model correctly selected all correct options. This is the most rigorous method of measuring multi-answer accuracy. 
    \item Precision: Precision measures the proportion of selected options that are actually correct. It captures how "careful" the model is in selecting options.
        \begin{equation}
        \text{Precision} = \frac{1}{N} \sum_{i=1}^{N} \frac{|\hat{Y}_i \cap Y_i|}{|\hat{Y}_i|}
        \label{eq:precision}   
        \end{equation}
    \item Recall: Recall measures the proportion of ground-truth correct options that are successfully selected by the model. It reflects the model's ability to cover all relevant answers.
        \begin{equation}
        \text{Recall} = \frac{1}{N} \sum_{i=1}^{N} \frac{|\hat{Y}_i \cap Y_i|}{|Y_i|}
        \label{eq:recall}
        \end{equation}
    \item JI(Jaccard Index): The Jaccard Index quantifies the similarity between the predicted and ground-truth answer sets by computing the size of their intersection over the size of their union.
        \begin{equation}
        \text{JI} = \frac{1}{N} \sum_{i=1}^{N} \frac{| \hat{Y}_i \cap Y_i |}{| \hat{Y}_i \cup Y_i |}
        \label{eq:ji}
        \end{equation}
\end{itemize}
\subsection{Count Bias Metrics}
These metrics measure the model’s ability to estimate the number of correct answers per question, regardless of which specific options are selected.
\begin{itemize}
    \item Mean Count Difference (CtDif):The average signed difference between the number of predicted and actual correct answers. A positive value indicates over-selection; a negative value indicates under-selection.
        \begin{equation}
        \text{CtDif} = \frac{1}{N} \sum_{i=1}^{N} \left( |\hat{Y}_i| - |Y_i| \right)
        \label{eq:placeholder}
        \end{equation}
    \item Mean Absolute Count Difference (CtDifAbs): The average absolute error between the predicted and true number of correct answers.
        \begin{equation}
        \text{CtDifAbs} = \frac{1}{N} \sum_{i=1}^{N} | \hat{Y}_i| - \left| Y_i \right|
        \end{equation}
    \item Count Accuracy (CtAcc): The proportion of questions for which the model predicted the exact number of correct answers, regardless of which specific options were selected.
        \begin{equation}
            \text{CtAcc} = \frac{1}{N} \sum_{i=1}^{N} 1 \left[| \hat{Y}_i| = \left| Y_i \right| \right]
        \end{equation}
\end{itemize}

\section{Human Refinement Procedure for Question-Answer Problem set Construction}
After the initial generation of QA sets using GPT-4o, all questions and answer options undergo a comprehensive human refinement process to ensure both quality and alignment with the visual data. The refinement process is guided by the following key principles:
\begin{itemize}
    \item Implicit Expression of Visual Information: If any numerical values or visual trends (e.g., increasing or decreasing patterns) present in the infographic are directly stated in the question or answer choices, they are rewritten into more implicit forms. This prevents the model from solving the question based solely on the text without engaging in visual interpretation.
    \item Correction of Mismatched or Ill-formed Questions: When the generated question is found to be factually inconsistent with the infographic or logically incoherent, it is manually rewritten. This includes regenerating the entire question and its answer choices to ensure alignment with the visual content.
    \item Causality-Type Consistency in Task 2: In Task 2, special attention is given to whether each question aligns correctly with its designated causality type (e.g., distinguishing between “Cause” and “Effect” questions). Misclassified cases are corrected to reflect the intended reasoning structure.
\end{itemize}
This human refinement process ensures that the resulting questions require genuine causal reasoning based on the infographic. By eliminating surface-level textual cues and enforcing logical and semantic coherence, we aim to construct a benchmark that meaningfully assesses a model’s ability to perform visual-grounded causal inference.

\begin{table*}[t]
\centering  
\small
\begin{tabular}{l|c|l|c|l|c|l|c}
\toprule
\multicolumn{8}{c}{\textbf{Distribution of Infographics}}\\
\midrule
Anatomical Illustration & 1 & Narrative Infographic & 1 & Scatter Plot & 21 & Bar Chart & 128 \\
\midrule
Mixed Bar and Pie Chart & 1 & Choropleth Map & 16 &Statistical Chart & 3  & Donut Chart & 2 \\
\midrule
Partial Geographical Infographic & 10 & Simple Infographic & 5 &Sankey Diagram & 3  & Heatmap & 1 \\
\midrule
World-wide Geographical Infographic & 9 & Simple Infographic & 5 & Statistical Graph & 2 & Area Chart & 1 \\
\midrule
Horizontal Bar Chart & 2 & Comparative Chart & 2 & Bubble Chart & 1 & Line Chart & 215\\
\midrule
Mixed Bar and Line Chart & 1 & Stacked Area Chart & 27 & Stacked Bar Chart & 17 & Pie Chart & 5\\
\midrule
Informational Infographic & 2 & Statistical Infographic & 9 & Thematic Map & 3 & Table  & 1\\
\midrule
Timeline Infographic & 1 &Vertical Bar Chart & 1  & Venn Diagram & 1 & Treemap & 1\\
\bottomrule
\end{tabular}
\caption{Distribution of Infographics in \benchmark}
\label{tab:full_infographic_distribution}
\end{table*}

\end{document}